%% file: PaperForReview.tex
\documentclass[10pt,twocolumn,letterpaper]{article}

\usepackage[pagenumbers]{cvpr} 

\usepackage{graphicx}
\usepackage{amsmath}
\usepackage{amssymb}
\usepackage{booktabs}
\usepackage{times}
\usepackage{mathtools}
\usepackage[normalem]{ulem}
\usepackage{enumitem}
\usepackage{bm}
\usepackage{framed,multirow}
\usepackage{float}
\usepackage{subcaption}
\usepackage{changepage}
\usepackage{cuted}
\usepackage{capt-of}

\usepackage{graphicx}
\usepackage{float}

\DeclareMathAlphabet{\mathcal}{OMS}{cmsy}{m}{n}

\DeclarePairedDelimiterX{\norm}[1]{\lVert}{\rVert}{#1}

\usepackage[pagebackref,breaklinks,colorlinks]{hyperref}

\usepackage[capitalize]{cleveref}
\crefname{section}{Sec.}{Secs.}
\Crefname{section}{Section}{Sections}
\Crefname{table}{Table}{Tables}
\crefname{table}{Tab.}{Tabs.}

\def\cvprPaperID{11119} 
\def\confName{CVPR}
\def\confYear{2022}

\begin{document}
\title{Depth Refinement for Improved Stereo Reconstruction}

\author{Amit Bracha \and Noam Rotstein\and David Bensaïd\and Ron Slossberg\and Ron Kimmel \\
Technion - Israel Institute of Technology\\
\tt\small\{amit.bracha, snoamr, dben-said, ronslos, ron\}@cs.technion.ac.il
}
\maketitle


\begin{figure*}[!h]
    \centering
    \includegraphics[width=0.9\textwidth]{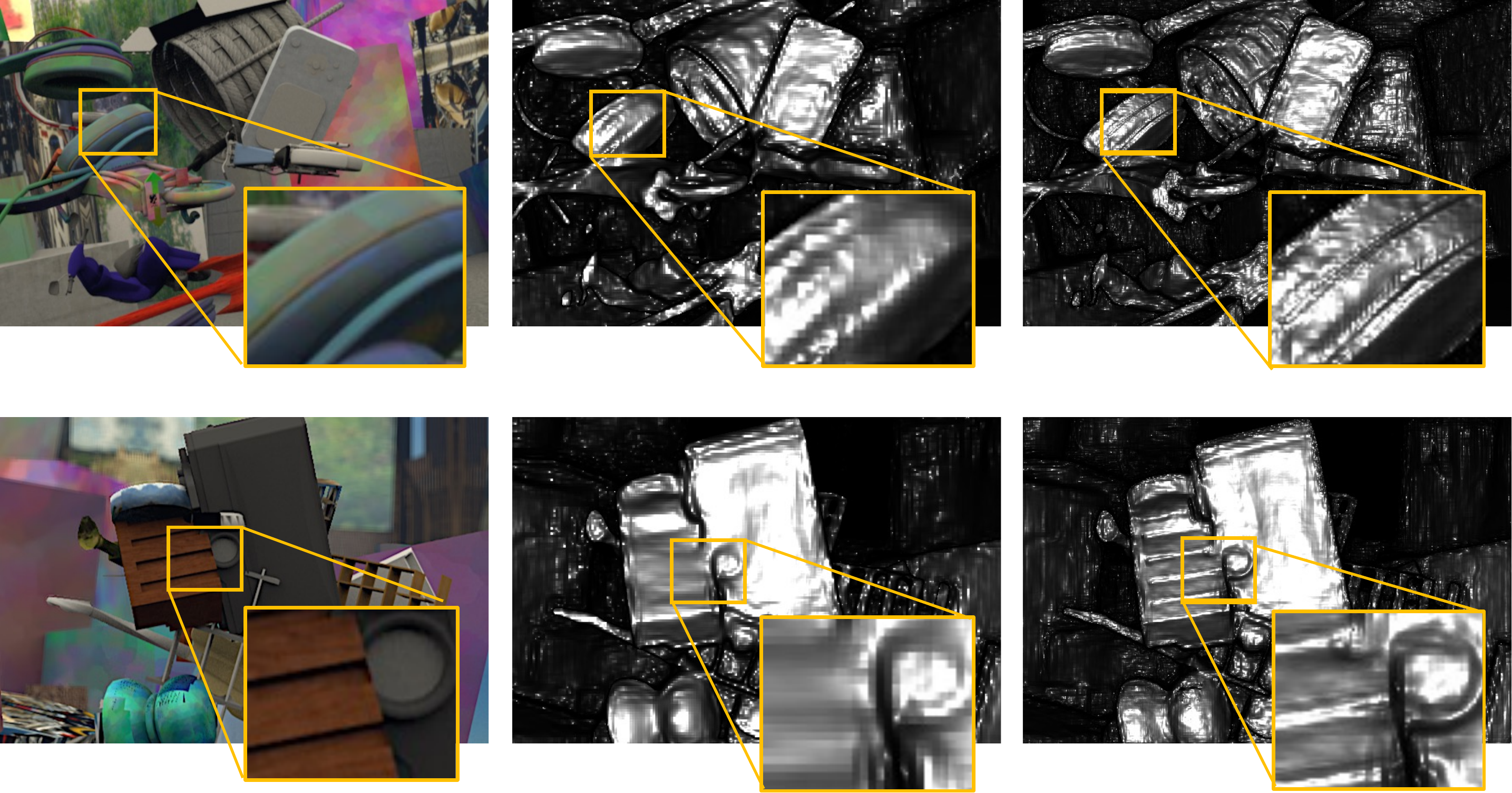}
    \caption{
    A selection of shading images ${(\left\| \nabla Z  \right\|+\epsilon)}^{-1}$ from the Sceneflow dataset \cite{sceneflow}.
     Left image inputs (left), shading images produced by the LEAStereo \cite{LEAStereo} depth estimation (middle) and shading images of the refined depth after applying our refinement process (right).
    We highlight areas demonstrating that the proposed refinement improves both edges as well as fine details within the depth image.
    }
    \label{fig:syntheticResults}
\end{figure*}

\begin{abstract}
Depth estimation is a cornerstone of a vast number of applications requiring 3D assessment of the environment, such as robotics, augmented reality, and autonomous driving to name a few. 
One prominent technique for depth estimation is stereo matching which has several advantages: it is considered more accessible than other depth-sensing technologies, can produce dense depth estimates in real-time, and has benefited greatly from the advances of deep learning in recent years. 
However, current techniques for depth estimation from stereoscopic images still suffer from a built-in drawback. 
To reconstruct depth, a stereo matching algorithm first estimates the disparity map between the left and right images before applying a geometric triangulation. 
A simple analysis reveals that the depth error is quadratically proportional to the object's distance. Therefore, constant disparity errors are translated to large depth errors for objects far from the camera. 
To mitigate this quadratic relation, we propose a simple but effective method that uses a refinement network for depth estimation. 
We show analytical and empirical results suggesting that the proposed learning procedure reduces this quadratic relation.
We evaluate the proposed refinement procedure on well-known benchmarks and datasets, like Sceneflow and KITTI datasets, and demonstrate significant improvements in the depth accuracy metric.
\end{abstract}


\input{intro}
\input{related_work}
\input{method}
\input{Experiments}
\input{Discussions}

{\small
\bibliographystyle{ieee_fullname}
\bibliography{egbib}
}

\end{document}

%% file: intro.tex
\section{Introduction}
\label{sec:intro}
Depth estimation is a well-studied task in computer vision with a central role in increasingly important areas such as robot navigation, augmented reality, and autonomous driving. 
Depth sensing can be achieved either by actively illuminating the environment and measuring various aspects of the reflected light or by passively sensing the ambient light from one or more viewpoints. 
The former method is considered more accurate and complex laser-based systems, such as LiDAR, are the gold standard in autonomous driving. 
However, the high cost of modern LiDAR systems as well as their limitations in real-time configurations restraint their wide application.
Stereoscopic depth estimation relies on estimating the relative displacement of an object between two views. 
This displacement is termed disparity and is inversely proportional to the object's distance. 
For this purpose, a dense correspondence between pairs of images is performed, followed by a disparity refinement phase.

In essence, stereo reconstruction trades hardware for software complexity, as the correspondence task is particularly challenging within textureless, poorly illuminated, and occluded regions. 
However, human high depth estimation capabilities with a natural stereo configuration suggest that a simple two-camera setup is sufficient for accurate spatial reasoning.
In recent years, stereo correspondence methods have greatly benefited from modern deep learning techniques and their quality is constantly improving. 
Current state-of-the-art methods achieve a $3$-pixel outlier rate of less than $2\%$ on standard benchmarks such as Middlebury, ETH3D, or Kitti \cite{middlebury, ETH3D,kitti2015}.

Despite the rapid improvements in the quality of stereo matching, range-from-stereo pipelines based on disparity estimation suffer from an inherent flaw.
As detailed in Section \ref{subsection: depth vs. disparity}, the depth error is \textit{quadratically} proportional to the depth of the observed object; for a constant error in disparity, the perceived error in depth is much larger for distant objects.
The previous quadratic relation is a major drawback of the stereo methods, making them unsuitable for applications such as autonomous driving where accurate distance estimation of far objects is essential.
In PseudoLidar++ \cite{pseudolidar++}, You et al. have pointed out this fundamental problem and addressed it by directly optimizing a model based on PSMNet \cite{PSMNet} on depth instead of disparity.
However, this approach also suffers from a significant drawback.
Modern range-from-stereo models, including PSMNet, are hinging on stereo matching algorithms where the disparity is the natural metric to optimize.
The transition to a direct depth-based approach negatively affects the performances of the model.
We propose a novel two-step approach to resolve the dichotomy between the essential metric for applications, depth, and the natural metric for the optimization of stereo matching algorithms, disparity. The two main steps comprising our methodology are the following:
Firstly, state-of-the-art stereo range estimation models are trained on disparity, as usual, to maximize the performance of the stereo matching algorithm. 
Secondly, a refinement model is depth-wisely-optimised using the computed disparity map and the left and right images.
A comparison between unrefined and refined depth estimates is depicted in Fig.~\ref{fig:syntheticResults}. We observe that the refined depth depicts finer detail and sharper object boundaries.

Our main contributions are the following:
\begin{itemize}
    \item We propose a novel approach to improve the accuracy of stereoscopic depth estimation.
    This procedure can easily be implemented on top of any existing depth-from-stereo model. 
    \item We demonstrate a significant improvement for several state-of-the-art methods by recasting the classical disparity refinement stage into a depth refinement problem.
    \item We present an ablation study demonstrating a clear advantage of our two-step refinement approach over frameworks trained end-to-end with a depth-based loss.
    \item We successfully leverage analytical relations between disparity and depth errors to efficiently correct discrepancies in-depth estimation.
\end{itemize}

%% file: related_work.tex
\section{Related work}
\subsection{Deep stereo matching}
The problem of finding the stereo correspondence between a pair of rectified images is well studied in the field of computer vision. 
In recent years, this task has benefited from the introduction of modern neural networks specifically adapted for the stereo correspondence task and is now largely dominated by such data-driven rather than classical model-driven methods.

The first to tackle stereo correspondence in the era of DNN were \cite{zbontar2016stereo} who proposed a patch matching scheme based on a triplet hinge loss fed by matching and non-matching examples. The learned patch descriptors provide the initial disparity estimates which are further refined using classical methods.
This approach, however, is somewhat limited due to its local nature focusing on the matching scheme while disregarding the global geometric structure. 
This was remedied by later efforts \cite{flownet}, who provide end-to-end frameworks for matching as well as disparity refinement.
In a later publication by \cite{GCNet}, termed GC-Net, 3D convolutions kernels were applied to the feature correspondence space, facilitating a more general framework for end-to-end correspondence and disparity refinement. 
This new approach provided a significant improvement over previous ones.

In the network proposed by \cite{PSMNet} termed PSMnet, the authors further improve the feature extraction with a multi-scale feature extraction network \cite{FPN} typically used for object detection.
In LEAStereo \cite{LEAStereo}, a neural architecture search is performed to further streamline the matching pipeline.
Another notable improvement was GA-net \cite{ganet}. 
There, an architecture inspired by the semi-global-matching (SGM) algorithm \cite{SGM} was incorporated into the deep stereo learning pipeline.
This is achieved by introducing semi-global as well as local guided aggregation layers which mimic the classical SGM method with tunable weights.

The work perhaps closest to our line of thought is \cite{pseudolidar++} who argue that depth, rather than disparity, is a more suitable metric for training and evaluation of geometry from stereo pipelines as the desired output from the model is depth estimation and not disparity. 
To this end, they suggest altering the model output to calculate the depth from the obtained disparity and adapt the cost aggregation layer accordingly.
However, their adjustment is not based on an axiomatic method, and it does not have any geometric foundations as the stereo matching algorithm.
Here, we propose to refine the output obtained from the stereo matching algorithm under metrics that evaluate the estimated depth rather than disparity. 
This subtle yet important difference allows us to retain geometric fidelity and at the same time significantly improve the more meaningful depth estimation metrics.

\subsection{Refinement for stereo matching}
In several recent papers, a refinement component has been introduced in addition to an existing pipeline to improve the disparity estimation quality.
For example, in \cite{pang2017cascade}, the authors utilize a cascade residual learning model as a refinement step applied to DispNet \cite{dispnet}.
The MSMD-Net \cite{msmdnet} proposed a multi-scale approach where left-right pairs were processed by a multi-resolution pyramid network and then refined, reducing inference complexity and subsequently run-time.
In a recent paper \cite{raftstereo}, the RAFT-stereo architecture was proposed. 
This model consists of a stereo estimation network coupled with a GRU-based multi-scaling refinement stage. 
All models mentioned in this section emphasize the refinement step of the matching process; however, they all address the problem as a disparity refinement process. 
Hence, these methods still suffer from the quadratic relation between the distance from the observed object and the depth error.

\subsection{Monocular depth estimation}
Monocular depth estimation methods attempt to estimate depth from a single image. 
Although this problem is ill-posed, by obtaining a good prior knowledge of the surrounding scenes and objects it is possible to obtain reasonable results. Such methods were previously proposed by \cite{qiao2021vip,lee2021patch,aich2021bidirectional,lee2019big,fu2018deep,vianney2019refinedmpl,xu2021pyramid} to name a few. 
Here, we propose to combine elements from both monocular and stereo-reconstruction to derive our depth refinement scheme.

In \cite{fu2018deep}, the authors incorporate a multi-scale network architecture coupled with an ordinal training loss. 
This loss effectively reduces the errors stemming from large depth values during training.
In addition, the authors of \cite{lee2019big} utilized an encoder-decoder architecture with an additional local layer as a refinement stage.
Furthermore, in \cite{lee2019big} the authors propose an end-to-end object-detection from a stereo model similar to \cite{pseudolidar++} supplemented by an additional refinement stage improving previous results obtained by similar approaches.

In contrast to monocular depth estimation schemes which solve an ill-posed problem, we retain geometric information captured by the stereo-matching pipeline and utilize it to better estimate the depth from the viewpoint of one or both of the stereo cameras. 
In addition, our goal is to refine depth estimates of existing models, hence, residual learning is more appropriate in our case than previously discussed monocular reconstruction losses.

%% file: method.tex
\section{Depth Refinement for Stereo Reconstruction}
Typical pipelines for depth reconstruction from stereoscopic image pairs generally rely on finding correspondences between the left and right images and are optimized according to a disparity criterion.
However, such optimization is error-prone for more distant objects and leads to a distorted reconstruction of the geometry, see Section \ref{subsection: depth vs. disparity}.
Here, we propose to mitigate this bias by introducing a refinement network trained to minimize a depth-based loss.

\subsection {Depth and disparity}
\label{subsection: depth vs. disparity}
The relation between depth and disparity is given by a simple geometric triangulation \cite{cormack1985computation},
\begin{eqnarray}
 Z_a &=& \frac{b\cdot f_x}{d_a}\, \propto\, \frac{1}{d_a}.
\label{disparity2depth}
\end{eqnarray}
Where $Z_a$ is the depth estimate, $b$ is the baseline of the camera, $f_x$ is the horizontal focal length, and $d_a$ is the disparity approximation.
Denote $d_{gt}$ the disparity ground truth, $\epsilon_d$ the error in disparity and $\epsilon_Z$ the error in depth.
Assuming $d_{gt} \gg \epsilon_d$, $\epsilon_Z$ can be approximated by,
\begin{eqnarray} 
\epsilon_Z \,=\, Z_{gt} - Z_a &= &\dfrac{bf_x}{d_{gt}} - \dfrac{bf_x}{d_a} \cr
& =&\dfrac{bf_x\epsilon_d}{d_{gt}(d_{gt} + \epsilon_d)} \cr
& \approx & \dfrac{bf_x\epsilon_d}{d_{gt}^2} \, = \, \dfrac{\epsilon_d }{bf_x}Z_{gt}^2.
\label{eq:error_disp2depth}
\end{eqnarray}
The above equation reveals that the quadratic relation between the error in depth and the distance of the observed object mentioned in Sec. \ref{sec:intro}.
Fortunately, this relation only holds as long as the range is determined by the disparity.
In PseudoLidar++ \cite{pseudolidar++}, You \etal promote training directly for depth estimation while skipping the initial correspondence stage.
On one hand, by abandoning the classical stereo matching paradigm, we can avoid the problem exposed in Eq. \ref{eq:error_disp2depth}.
On the other hand, reconstructing depth from disparity is a natural approach with geometric foundations. 

Here, we propose a new method for processing depth estimation while also preserving the geometric guarantees stemming from correspondence and triangulation.
In contrast to the end-to-end approach proposed by \cite{pseudolidar++}, we propose a two-step method that splits the overall pipeline into a disparity estimation stage followed by a dedicated depth refinement model.
The disparity approximation stage is obtained by adopting several successful stereo matching pipelines such as the ones discussed in the previous section. 
The initial disparity approximation is then converted into a depth estimation that is further refined via a dedicated DNN model, (see Sec. \ref{Multiplication_residual_refinement}).

\subsection{Multiplicative residual refinement} \label{Multiplication_residual_refinement}
Motivated by modern denoising procedures \cite{residual_denoising}, we consider a residual architecture.
Denote $Z$ the final depth estimate, $Z_e$ the estimate before refinement, $R$ the output of the refinement process, and $\oplus$ the pixel-wise addition. 
We have,
\begin{eqnarray}
    Z &=& Z_e \oplus R.
\end{eqnarray}
According to Eq.\ref{eq:error_disp2depth}, the refinement $R$ should be proportional to $Z^2$.
However, approximating a polynomial function of the input values by a neural network is not trivial \cite{Joseph-Rivlin_2019_ICCV,liang2019deep, yarotsky2017error}. 
Theoretically, representing  a polynomial function of the input requires deep architectures.
The network's complexity is bounded from below by the desired error and the degree of the polynomial \cite{liang2019deep, yarotsky2017error}.
Beyond simply increasing the number of parameters, the introduction of domain-specific prior knowledge typically has a positive influence on model performance.
A successful example of such utilization of prior knowledge in 3D Computer Vision is MomeNet \cite{Joseph-Rivlin_2019_ICCV}, a neural network dedicated to processing point clouds based on the previously introduced PointNet\cite{pointnet}.
MomeNet demonstrates improved performance over PointNet by exploiting the importance of moments in shape analysis. 
This is achieved by providing the model with the second-order moments of the points in addition to their coordinates.
Along the same line of thought, we leverage the analytical knowledge of the relation between the depth error and the distance to the observed object (Eq. \ref{eq:error_disp2depth}).
This is achieved by multiplying the model output by the depth estimate $Z_e$ instead of directly learning $R$.
\begin{eqnarray}
    R &=& f(Z_a,I_l,I_r;\theta)\odot Z_a.
\end{eqnarray}
Where $\odot$ is the Hadamard product and $f(Z_a,I_l,I_r;\theta)$ is the refinement network with weights $\theta$.
Multiplying the network's output by the depth estimates allows the network to easily converge towards a $R$ proportional to $Z^2$.
We quantitatively show the benefit of the former operation in Sec. \ref{subsec: ablations}.
\subsection{Depth Refinement (DR) Network}
The refinement model proposed is a CNN based on U-net \cite{unet} that learns depth priors from images to ameliorate the existing depth map.
The inputs to the DR network are the depth estimate computed by pretrained stereo matching pipeline, the left image, and the warping \cite{pang2017cascade} of the right image with respect to the disparity approximation.
That is,
\begin{eqnarray}
    R &=& f(Z_a,I_l,I_r^l;\theta)\odot Z_a,
\end{eqnarray}
and the output of our method is,
\begin{eqnarray}
    Z &=& Z_a \oplus (f(Z_a,I_l,I_r^l;\theta)\odot Z_a).
\end{eqnarray}
Where $I_r^l$ is the right image warped into the left image plane.
As previously discussed, we utilize a depth based loss during the training process.
\begin{eqnarray}
\label{depth loss}
    \mathcal{L} &=& \norm[]{Z_{gt} - Z_a \oplus f(Z_a,I_l,I_r^l;\theta)\odot Z_a}_1,
\end{eqnarray}
For numerical stability, the DR network is trained with a slightly modified version of the loss presented in Eq. (\ref{depth loss}),
\begin{eqnarray}
\mathcal{L} &=& \left \|\frac{Z_{gt} - Z_a}{Z_a} - f(Z_a,I_l,I_r^l;\theta)\right \|_1.
\end{eqnarray}
This modified loss produces range invariant gradients.
To improve the numerical stability during training, we ignore objects at a distance greater than $100m$.
This assumption is often applied in standard benchmarks such as KITTI to assess results.

%% file: Experiments.tex
\begin{table*}[h!]
\begin{center}
\begin{tabular}{|c|| c c | c  c | c  c | c c |} 
 \hline
 \multirow{2}{*}{Methods} & \multicolumn{2}{c|}{Depth error} & \multicolumn{2}{c|}{EPE} & \multicolumn{2}{c|}{D1 - 1px} & \multicolumn{2}{c|}{D1 - 3px} \\ 
 & with DR & w/o DR & with DR & w/o DR & with DR & w/o DR & with DR & w/o DR \\
 \hline
 LEASereo \cite{LEAStereo}& 0.596 & 1.092 & 0.810 & 0.861 & 8.1\% & 8.8\% & 3.2\% & 3.2\% \\ 
 
 RAFT-Stereo \cite{raftstereo} & 0.582 & 0.744 & 1.034 & 1.047 & 11.1\% & 11.1\% & 5.7\% & 5.8\% \\
 
 CFNet \cite{cfnet} & 0.734 & 1.12 & 1.025 & 1.173 & 10.9\% & 12.29\% & 5.4\% & 6.6\% \\
 \hline
\end{tabular}
\end{center}
\caption{An evaluation of our model on the Sceneflow\cite{sceneflow} test-set.
We examine our method combined with three stereo matching pipelines.
The results demonstrate that our method (DR) outperform reported accuracy in all the metrics with the largest improvement in the depth estimation error. *Reported results might differ with respect to published figures do to different masking protocols.
}
\label{table:Sceneflow evaluates}
\end{table*}


\section{Experiments}
In this section, we evaluate the proposed refinement procedure by applying it to three recent stereo matching pipelines, the CfNet, the LEA-Stereo, and the RAFT-Stereo.
Experiments are performed on two classical benchmarks: the synthetic dataset Sceneflow \cite{sceneflow} and KITTI 2015 \cite{kitti2015}.
We demonstrate that the proposed refinement methodology substantially improves the depth estimate of each model, see tables~\ref{table:KIITI2015_evaluates} and~\ref{table:Sceneflow evaluates}. 
In addition, we observe an improvement in the performance of the studied stereo methodologies in several common disparity-related metrics as well.
We performed an ablation study in which we examine a similar architecture and training procedure but vary the loss function between an additive or multiplicative residual loss, see Table~\ref{table:Additive_vs_multiplication}. 
Finally, we compare our approach to PseudoLidar++ \cite{pseudolidar++}, see Table~\ref{table:DR_vs_SDN}, which proposes a line of thought close to ours and also focus on depth.

\subsection{Datasets}
\paragraph{Sceneflow.}
As previously discussed, we evaluate the proposed refinement pipeline on two well know datasets, Sceneflow and KITTI 2015.
Sceneflow is a synthetic dataset presented in \cite{sceneflow}, containing roughly 35K images for training and 4.3K images as a test-set.
The main benefit of such synthetic datasets is their dense and precise ground truth and their large size making them suitable for the training of DNNs.
On the other hand, these types of datasets present unique challenges when transferring trained models to real domains due to many factors such as sensor noise, lens deformations, scene composition, and realism.
Example scenes from the dataset as well as their estimated and refined depth outputs are depicted in Fig.~\ref{fig:syntheticResults}.
Nevertheless, we utilize this dataset to achieve several goals.
(i) We demonstrate our ablation experiment using this data. (ii) We examine our model performance without the confounding factors of ground truth noise, misalignment, miscalibration, etc.
(iii) We utilize its vast size to perform transfer learning to the smaller real KITTI dataset.

\paragraph{KITTI 2015.}
To evaluate our model on real-world data, we turn to the KITTI 2015 dataset \cite{kitti2015}.
This widely used benchmark includes a relatively small training-set containing $200$ images paired with sparse ground-truth depth maps as a similarly sized test-set with held-out ground-truth.
The KITTI dataset was obtained by mounting cameras as well as a Velodyne LIDAR onto a car and driving it on roads and city streets while recording.
The LIDAR was calibrated and synchronized with the stereo cameras thus producing the corresponding depth to each dataset image.
The depth sparsity stems from the scanning laser technology employed by the Velodyne scanner.
The sparse nature of the ground-truth images creates some difficulty when evaluating highly accurate stereo correspondence pipelines as the ground-truth is typically inaccurate near object boundaries which are often the differentiating factor between the various tested methods.
Therefore, methods that perform well at object boundaries and may receive high accuracy scores on benchmarks containing dense depth maps might under-perform on the KITTI benchmark. 
We note that the proposed refinement process based on depth and image data is especially successful at object boundaries and therefore suffers from the sparse nature of the KITTI ground-truth depth during evaluation.

\begin{table*}[h!]
\begin{center}
\begin{tabular}{|c|| c  c | c  c |} 
\hline
\multirow{2}{*}{Methods} & \multicolumn{2}{c|}{Depth error} & \multicolumn{2}{c|}{EPE} \\ 
 & with DR & w/o DR & with DR & w/o DR \\
 \hline
 LEASereo &  \multicolumn{2}{c|}{$0.43 \pm 0.03$  \, $0.47 \pm 0.04$} & \multicolumn{2}{c|}{$0.49 \pm 0.04 $ \, $0.50 \pm 0.04$}  \\ 
 
 RAFT-Stereo & \multicolumn{2}{c|}{$0.44 \pm 0.03 $ \, $0.47 \pm 0.03$ }& \multicolumn{2}{c|}{$0.49 \pm 0.04$ \, $0.49 \pm 0.04$} \\
 
 CFNet &  \multicolumn{2}{c|}{$0.46 \pm 0.03$  \,  $0.50 \pm 0.03$ }&\multicolumn{2}{c|}{ $0.57 \pm 0.05$ \, $0.58 \pm 0.04$} \\
 \hline
\end{tabular}
\end{center}
\end{table*}

\begin{table*}[h!]
\begin{center}
\begin{tabular}{|c|| c  c | c  c | c  c | c c|} 
 \hline
 \multirow{2}{*}{Methods} & \multicolumn{2}{c|}{D1 - 1px} & \multicolumn{2}{c|}{D1 - 3px} \\ 
  & with DR & w/o DR & with DR & w/o DR \\
 \hline
 LEASereo & \multicolumn{2}{c|}{$8.6 \pm 1.1$\% \, $9.0 \pm 1.3$ \%} & \multicolumn{2}{c|}{$1.2 \pm 0.2 $ \% \, $1.1 \pm 0.1 $ \% }\\ 
 
 RAFT-Stereo &  \multicolumn{2}{c|}{$8.7 \pm 1.3 \%$ \, $8.7 \pm 1.2 \% $ }& \multicolumn{2}{c|}{$1.4 \pm 0.3 \% $ \, $ 1.3 \pm 0.3 \% $} \\
 
 CFNet & \multicolumn{2}{c|}{$10 \pm 1.5 \% $ \, $11 \pm 1.5 \% $} & \multicolumn{2}{c|}{$ 1.7 \pm 0.5 \% $ \, $ 1.6 \pm 0.3 \%  $}\\
 \hline
\end{tabular}
\end{center}
\caption{Evaluation of our model on the KITTI 2015 dataset \cite{kitti2015}.
To evaluate our model on the training-set, we used a $10$-fold validation scheme.
We examine our method on three different stereo matching algorithms, where LEAStereo \cite{LEAStereo} is the first ranked algorithm on KITTI 2015 among methods with published code.
The results demonstrate a performance increase of roughly $7\%$ in depth estimation over the baseline.
We observe that the EPE metric, the disparity error, and D1-1px, the percentage of pixels with more than 1px error in disparity, are also improved.}
\label{table:KIITI2015_evaluates}
\end{table*}

\subsection{Depth Refinement Evaluation}
We perform an extensive evaluation of the proposed depth refinement methodology using the aforementioned datasets.
First, we evaluate our approach on the synthetic SceneFlow dataset. 
We conduct our evaluation on three recent stereo correspondence pipelines, namely: LEAStereo~\cite{LEAStereo}, CFNet~\cite{cfnet}, and RAFT -Stereo~\cite{raftstereo}.
To perform a fair evaluation, we load the author-provided weights for each model trained specifically for each dataset, and use the same data preprocessing procedures appropriate for each method.

We test the refined depth output under both depth as well as disparity-related metrics. 
Since our refinement scheme is trained according to a depth $L_1$ loss, it is, therefore, possible that the increase in depth accuracy is achieved at the cost of disparity degradation.
To evaluate the disparity, we transform the refined depth estimate back to disparity through
\begin{equation}
    d_a = \frac{b\cdot f_x}{Z},
\end{equation}
where $d_a$ is the refined disparity estimate.
All the results are evaluated only in unmasked areas where the disparity is not greater than the max-disparity hyper-parameter of the model, and the depth is not greater than $100$ meters.
The obtained refined depth and disparity errors are presented in Tab. \ref{table:Sceneflow evaluates}. 
We find that our refined depth estimates demonstrate a $15\%$ to $39\%$ improvement in depth accuracy over the baseline.

Furthermore, the refined disparity derived from the estimated depth is improved as well relative to the baseline, even though the refinement process does not aim to directly improve disparity errors.
These results demonstrate that while our methodology achieves lower depth error rates as expected, the disparity estimation does not suffer and even manages to benefit from it as well.

To examine our method on the KITTI 2015 dataset, we use the same three deep learning matching pipelines used with Sceneflow and load the appropriate pretrained weights specifically used for KITTI.
To facilitate network learning and avoid over-fitting to the small dataset, we employ transfer learning.
Since the ground-truth of the test is not released and the depth values are unknown, we evaluate our refinement quality via $10$-fold cross-validation on the training-set. 
We start by training on the Sceneflow dataset and then progress to training only on a specific data fold of the KITTI dataset. 
Each fold is then trained for $1500$ additional epochs.
The result in Tab. \ref{table:KIITI2015_evaluates}, shows that our method successfully ameliorates the depth estimate, and it reduces the error by about $7\%$.

\subsection{Ablation Study}
\label{subsec: ablations}
\paragraph{Benefits of the two-step pipeline.} To evaluate the benefit of performing a two-step refinement in contrast to end-to-end learning of depth estimation, we compare our results to the single-step SDN model \cite{pseudolidar++}.
In \cite{pseudolidar++}, the authors used as a backbone to their network, the PSMNet model \cite{PSMNet}, and replace the disparity-based training loss with a depth estimation loss.
We compare the results of our proposed pipeline applied to PSMNet with the reported one-step training reported by \cite{pseudolidar++}.
We used the pretrained network weights from Sceneflow published by the authors for both the PSMNet as well as SDN.
We trained our refinement scheme for $20$ epochs on the Sceneflow dataset and present the result in Tab.~\ref{table:DR_vs_SDN}.
The results demonstrate that our proposed method outperforms both the PSMNet baseline and the version trained on depth as reported by \cite{pseudolidar++}.
In addition, we observe that our method with the PSMNet reached a lower error in depth in comparison to the SDN model.

\begin{table}[h!]
\begin{center}
\begin{tabular}{|c|| c | c | c|} 
 \hline
 Methods & Depth error & EPE & D1 \\ [0.5ex] 
 \hline
 PSMNet & 1.55 & 1.99 & 7\% \\ 
 PSMNet + DR & 0.582 & \textbf{1.31} &\textbf{6.7\% }\\
 SDN & 0.641 & 2.38 & 9.7\% \\
 SDN + DR  & \textbf{0.58} & 1.89 & 8.7\% \\ 
 \hline
\end{tabular}
\end{center}
\caption{We conduct a comparison between two pretrained models: stereo depth network (SDN) \cite{pseudolidar++}, and PSMNet \cite{PSMNet}, with and without depth refinement (DR).
We measure average absolute depth error (Depth error), end-point error (EPE), and percentage of pixels with more than $3$ pixels of disparity error (D1). 
We observe that the proposed refinement improves all tested metrics with relation to the baseline.
Note that reported results might differ from published figures due to different masking protocols.
}
\label{table:DR_vs_SDN}
\end{table}

\paragraph{Naive model.}
To refine the depth estimate, the output of our model is multiplied by the depth estimate produced by the stereo matching algorithm.
In this part, we wish to examine this multiplicative residual approach in contrast to a more naive additive approach as discussed in Sec. \ref{Multiplication_residual_refinement}.
To this end, we test the refinement model with the naive additive residual approach. 
Instead of multiplying by the depth estimate of the stereo model, we simply add the refinement output to the depth estimate.
Denoting the DR network output $f(Z_a,I_l,I_r^l;\theta)$ and the pre-computed depth map $Z_a$, the multiplicative residual approach refers to the computation of the final depth estimates $Z$ as $Z = Z_a \oplus f(Z_a,I_l,I_r^l;\theta)\odot Z_a$. 
The naive refinement refers to the results obtained by opting for a straightforward approach where the refinement output of the DR network is directly added to the pre-computed depth map, i.e $Z = Z_a \oplus f(Z_a,I_l,I_r^l;\theta)$.
We reevaluate the former method with the three stereo correspondence pipelines: CFNet, LEAStereo, and RAFT-Stereo. 
We substitute only the residual component and retain all the training procedures as before while training on the Sceneflow train-set. 
We report the results on the Sceneflow test-set comparing the two approaches. 
As before, each model has been trained for $30$ epochs.
The results presented in Tab.~\ref{table:Additive_vs_multiplication} demonstrate that while both approaches improve depth estimation, the multiplicative approach outperforms the naive additive one. 
This aligns with our expectation as discussed in Sec.~\ref{Multiplication_residual_refinement}.

\begin{table}[h!]
\begin{adjustwidth}{-0.15cm}{}
\begin{center}
\begin{tabular}{|c|| c | c | c|} 
 \hline
 Methods & LEAStereo & RAFT-Stereo & CFNet \\
 \hline
 Ours & \textbf{39.7\%} &\textbf{ 15.5\%} & \textbf{22.5\%} \\ 
 Naive refinement & 33.9\% & 13.6\% &  18.2\% \\
 \hline
\end{tabular}
\end{center}
\caption{ The results of the ablation study regarding two styles of residual learning. 
Indeed, the multiplicative approach consistently outperforms the naive additive approach across all tested correspondence methods. 
This result supports our expectations regarding quadratic functions and neural networks.}
\label{table:Additive_vs_multiplication}
\end{adjustwidth}
\end{table}
\paragraph{Quadratic error mitigation.}
Our proposed method for depth refinement of stereo matching algorithms attempts to tackle and reduce the quadratic relation between the depth and the error within the depth estimate. 
In this experiment, we assess the reduction of the quadratic relation after refinement when compared to the unrefined depth error.

\begin{figure}[h]
    \centering
    \includegraphics[width=0.51\textwidth]{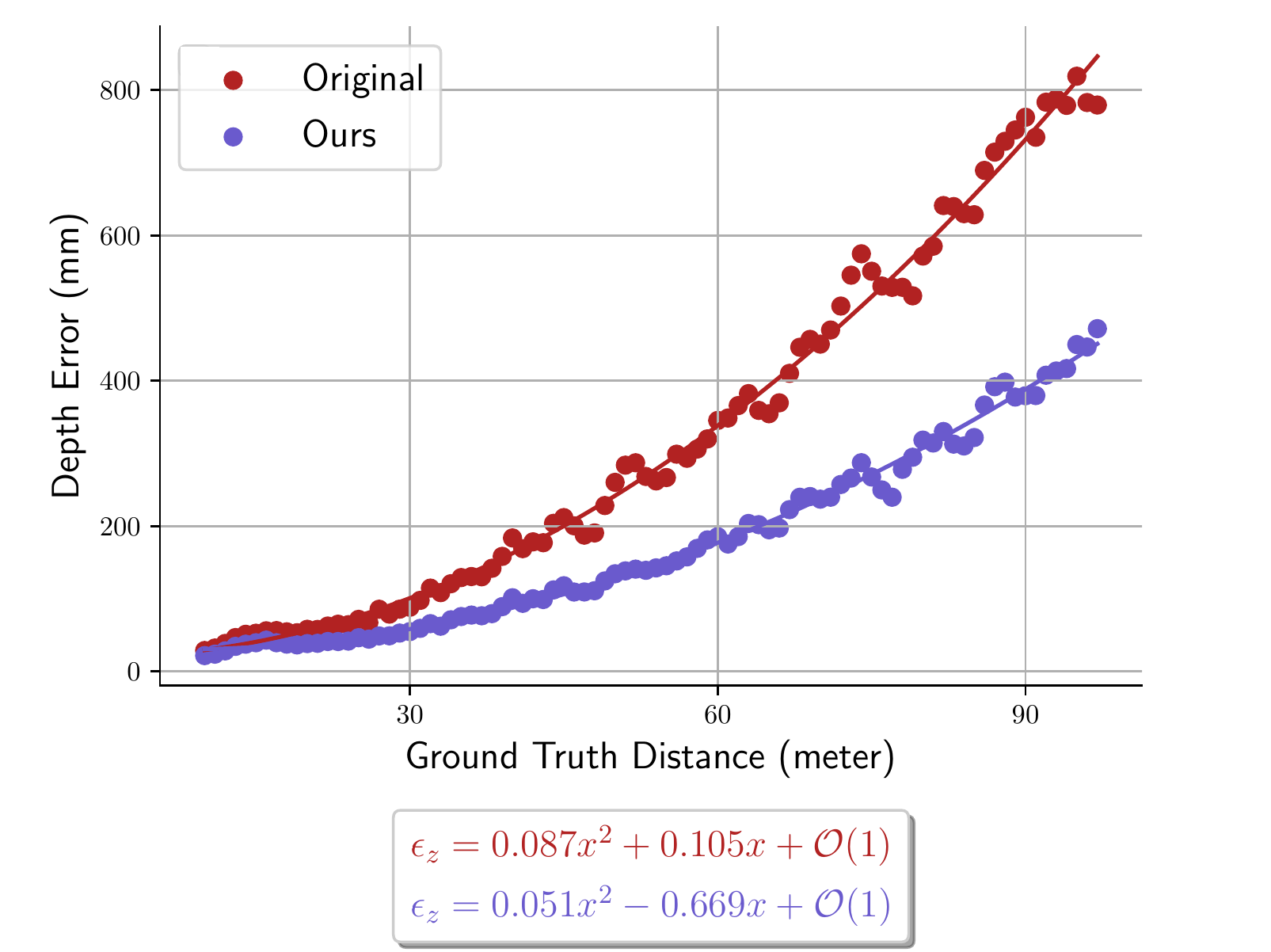}
    \caption{A comparison between the depth estimate of the baseline stereo algorithm and the refined estimate and fit a curve representing the quadratic relation of the error with the distance. 
    The x-axis is the ground-truth distance while the y-axis represents the median depth error in mm.
    By examining the coefficients of the fitted polynomials, we notice that the quadratic relation between the error and the distance is reduced due to the refinement process.
    }
    \label{fig:quadratic_relation}
\end{figure}

We first obtain the depth estimation resulting from the LEAStereo model evaluated on the Sceneflow test-set.
We measure the depth error with respect to the ground-truth with our refinement and without it.
To this end, we divided the pixels into bins according to their ground-truth distance with one meter intervals.
We then take the median depth error for each bin and plot the medians before and after the refinement process.
We present the result with a fitted second-degree polynomial trend line in Fig.~\ref{fig:quadratic_relation}.
The polynomial coefficients demonstrate a substantial reduction in the quadratic component when refining the depth estimate with our proposed method.

%% file: Discussions.tex
\section{Discussions and Conclusions}
In this paper, we demonstrate experimentally that processing the depth rather than the disparity, we can refine depth recovery for state-of-the-art depth from stereo pipelines. 
In addition, we find that the gained benefit in depth estimation quality is independent of the performance of the tested stereo pipeline and that even highly successful methods are amenable to our refinement approach.
Moreover, we demonstrate that, while our approach is geared towards improved depth estimation, the disparity-related metrics do not suffer a degradation in quality. 
This suggests that the depth domain, where natural geometric relations are preserved, is more tractable for learning via DNN's.

When attempting to refine disparity values, a DNN ideally models the relation between the disparity and the scene geometry; however, this relation is indirect as several transformations including homography transform, lens distortion, and depth triangulation are necessary mathematical transformations relating between perceived disparity and physical scene geometry.
Our ablation study demonstrates that even a simple residual denoising-based refinement model significantly improves depth estimation accuracy.
Additionally, it also improves disparity estimate fidelity, thus, validating the powerful notion of processing the model output in a more natural domain.
Furthermore, we show that multiplying the refinement model output by the input depth estimate before applying the loss function helps the model to converge in a shorter time, and improves the final estimation quality significantly.
This result aligns with our mathematical reasoning regarding the quadratic relation between the depth and the distance to the observed object as well as previous observations regarding the ability of networks to process quadratic functions.

Finally, we assess the relation between distance and depth error before and after our proposed refinement procedure. 
By doing so, we can demonstrate that the refined output exhibits a weaker quadratic relation.
The former points highlight that our framework is more robust to the object distance than previous disparity-based approaches.
Thus, it potentially opens the door to more complex and accurate depth-based refinement methodologies for stereo reconstruction, possibly leading to a linear relation in future work.